# PSO Fuzzy XGBoost Classifier Boosted with Neural Gas Features on EEG Signals in Emotion Recognition

Seyed Muhammad Hossein Mousavi
*Developer at Pars AI Company*
Tehran, Iran
hossein-mousavi@ieee.org and mosavi.a.i.buali@gmail.com

*Abstract*—Emotion recognition is the technology-driven process of identifying and categorizing human emotions from various data sources, such as facial expressions, voice patterns, body motion, and physiological signals, such as EEG. These physiological indicators, though rich in data, present challenges due to their complexity and variability, necessitating sophisticated feature selection and extraction methods. NGN, an unsupervised learning algorithm, effectively adapts to input spaces without predefined grid structures, improving feature extraction from physiological data. Furthermore, the incorporation of fuzzy logic enables the handling of fuzzy data by introducing reasoning that mimics human decision-making. The combination of PSO with XGBoost aids in optimizing model performance through efficient hyperparameter tuning and decision process optimization. This study explores the integration of Neural-Gas Network (NGN), XGBoost, Particle Swarm Optimization (PSO), and fuzzy logic to enhance emotion recognition using physiological signals. Our research addresses three critical questions concerning the improvement of XGBoost with PSO and fuzzy logic, NGN's effectiveness in feature selection, and the performance comparison of the PSO-fuzzy XGBoost classifier with standard benchmarks. Acquired results indicate that our methodologies enhance the accuracy of emotion recognition systems and outperform other feature selection techniques using the majority of classifiers, offering significant implications for both theoretical advancement and practical application in emotion recognition technology.

*Keywords*—*PSO; Fuzzy; XGBoost; Neural Gas Network (NGN); Feature Selection; EEG Signals; Emotion Recognition*

## I. INTRODUCTION

Emotion recognition is the process of identifying human emotion, typically from facial expressions, voice patterns, and physiological signals. This technology has significant applications across various fields, such as marketing, healthcare, automotive, psychology, and more [1, 2]. Physiological signals are measurable indicators that reflect the activity or condition of the human body, such as heart rate, brain waves (EEG), or blood pressure. These signals are commonly used in medical and psychological research to understand health, emotional responses, and bodily functions [3]. Physiological signals, such as heart rate, skin conductance, and EEG (electroencephalogram) measurements, provide a noninterfering means to assess an individual's emotional state [3]. These signals are crucial for emotion recognition as they offer a direct pathway to understanding the autonomic nervous responses that accompany different emotions. Integrating physiological signals into emotion recognition systems poses unique challenges, primarily due to the complexity and variability of the data. This complexity necessitates robust feature selection and extraction techniques to interpret the signals effectively. Traditional feature selection methods often struggle to handle high-dimensional data or data with complex nonlinear relationships.

The Neural Gas Network (NGN) [6], an advanced unsupervised learning algorithm, offers a compelling solution to these challenges. Unlike more traditional methods, NGN adapts to the input space without predefined grid structures, making it highly effective for complex feature extraction from physiological data. NGN has been applied successfully in areas like robotics for learning sensorimotor control [4], in finance for anomaly detection in transaction data [5], in health care for brain tumor segmentation [7], and more. The NGN is an unsupervised learning algorithm that can be used for feature extraction. It operates similarly to other competitive learning algorithms like Self-Organizing Maps (SOM) [8], but with a few key differences that affect how it learns the topology of the input space. NGN doesn't have a predefined grid structure as SOM does; instead, it uses a set of neurons that adaptively position themselves in the input space according to competitive learning rules. During training, input vectors are presented to the network, and the neurons compete to be the closest to these inputs. Over iterations, the neurons move closer to the input vectors, effectively learning to represent different regions of the input space. The final positions of the neurons after training can be used as features that capture the underlying structure of the data, making them useful for downstream tasks like classification or regression in classifiers like XGBoost. The XGBoost (Extreme Gradient Boosting) is an optimized distributed gradient boosting library designed to be highly efficient, flexible, and portable. It implements machine learning algorithms under the Gradient Boosting framework, providing a scalable, fast, and accurate method for regression, classification, and ranking problems.

However, despite advancements in feature selection via NGN, emotion recognition systems often face the challenge of optimizing model performance and handling fuzzy data. This is where fuzzy logic [9] becomes essential. Fuzzy logic allows the incorporation of reasoning that resembles human decision-making by handling degrees of truth rather than the usual binary true/false found in classical logic. Its applications range from consumer electronics, like air conditioners and washing machines [11], which adapt their operations to changing conditions, to advanced automotive systems [12], which improve vehicle control and stability, and in expert systems, when human expert assistance is needed [10], and more.

Moreover, the optimization [14] of complex models such as XGBoost, which is often used in emotion recognition for its effectiveness in classification tasks, can benefit greatly from any optimization algorithm such as Particle Swarm Optimization (PSO) [13]. PSO is a computational method used in artificial intelligence that optimizes a problem by iteratively trying to improve a candidate solution with regard to a given measure of quality. It simulates the social behavior of birds or fish to find the optimal solution in a search space. PSO helps fine-tune hyperparameters and optimize decision processes within XGBoost, which traditional optimization techniques may not efficiently solve.

In standard XGBoost classifiers, challenges like overfitting, handling non-linear relationships, and hyperparameter tuning are prevalent. By integrating PSO and fuzzy logic into XGBoost, we address these issues comprehensively, enhancing model accuracy and robustness. The PSO-fuzzy approach allows for a nuanced handling of uncertainty and optimization of the learning process, leading to improved performance in emotion recognition tasks.

We are looking to answer the following question: RQ1: Does improving XGB by PSO and Fuzzy techniques lead to better performance of the trained model regarding final test accuracy? RQ2: Can NGN effectively select the most impactful features from emotional and physiological signals compared to benchmark feature sections? RQ3: Is the PSO fuzzy XGB classifier capable of surpassing other benchmark classifiers, including normal XGB, regarding different metrics?

In our research, we have achieved significant results in emotion recognition using physiological signals by employing these advanced methodologies. Not only have we improved the test accuracy of emotion recognition, but NGN features also provide better performance compared with other feature selection techniques in general to the classifiers.

Section 2 pays to related works done by other researchers in the field of improving XGBoost in emotion recognition and also using NGN for different applications, especially in feature selection. Section 3 covers the theoretical background, section 4 covers our proposed method, and section 5 covers our validations and results. Finally, the conclusion packs up our contribution. You can find the GitHub repository of our contribution's implementation by Python in the footnote[1].

## II. RELATED WORKS

This section covers research related to improving the XGBoost classifier using optimization algorithms, fuzzy logic, and other methods. Also, we investigate different feature selection algorithms in the field of emotion recognition. Just in order to save space, all is included in Table 1.

## III. THEORETICAL BACKGROUND

### A. Fuzzy Logic to Enhance XGBoost

Fuzzy logic can enhance XGBoost by incorporating uncertainty and partial truth into the model's decision-making process. In traditional machine learning models, including XGBoost, inputs and outputs are crisp values. By integrating fuzzy logic, the model can handle more ambiguous data or situations where the boundaries between classes in the classification are not clear-cut. Fuzzy systems modify the structure of XGBoost by fuzzifying the input features or the decision criteria used in the trees, which allows for softer decision boundaries and potentially higher tolerance to noisy data or outliers. This integration can lead to better generalization over complex or imprecisely defined data sets.

$$\bar{Y} = Aggregate\ (FuzzyDecision\ (\mu(X))) \qquad (1)$$

In which, $\mu(X)$ represents the fuzzification of the input features $X$. FuzzyDecision encapsulates the fuzzified decision criteria used in the XGBoost trees. Aggregate denotes the aggregation of outputs from all the fuzzy decision trees to produce the final model output $\bar{Y}$.

### B. Particle Swarm Optimization (PSO) to to Improve XGBoost

When applied to enhance XGBoost, PSO can be used to optimize the hyperparameters of the model (like learning rate, max depth of trees, or min child weight) more effectively than traditional grid search or random search methods. PSO iteratively improves a candidate solution with regard to a given measure of quality (such as minimizing the loss function). Each particle in the swarm represents a potential solution and moves through the hyperparameter space by following the best-found positions in the search space, which are updated as better positions are discovered by the swarm. This can lead to finding a more optimal set of

---

hyperparameters for XGBoost, potentially improving the model's performance on the given task.

$$Optimal\ Hyperparameters\ = PSO(\ Loss, X, Y) \qquad (2)$$

TABLE I.    RELATED WORKS TABLE

| Author(s) | Subject | Contribution | Year | Cite |
|---|---|---|---|---|
| Jiang, Hui, et al | PSO XGBoost – No Emotion | Network intrusion detection based on PSO-XGBoost model | 2020 | [15] |
| Yu, Jingxin, et al | PSO XGBoost – No Emotion | PSO-XGBoost Model for Estimating Daily Reference Evapotranspiration in the Solar Greenhouse. | 2020 | [16] |
| Unnisa, Mahera, and V. Ganesan | XGBoost Optimization – Facial Expression | Improved XGBoost Classifier for Micro Expression Recognition using Hybrid Optimization Algorithm | 2024 | [17] |
| Asemi, Hanie, and Nacer Farajzadeh | GWO–XGBoost – EEG Emotion | Improving EEG Signal-based Emotion Recognition using a hybrid GWO-XGBoost Feature Selection Method | 2023 | [18] |
| Costache, Romulus, et al | Fuzzy XGBoost - No Emotion | Flash-flood potential index estimation using fuzzy logic combined with deep learning neural network, naïve Bayes, XGBoost and classification and regression tree | 2022 | [19] |
| Zong, Jing, et al | Fuzzy XGBoost – EEG Emotion | FCAN–XGBoost: a novel hybrid model for EEG emotion recognition | 2023 | [20] |
| Dhara, Trishita, Pawan Kumar Singh, and Mufti Mahmud | Fuzzy XGBoost – EEG Emotion | The fuzzy ensemble-based deep learning model for EEG-based emotion recognition | 2024 | [21] |
| Nawaz, Rab, et al | Carious Features selection – EEG Emotion | Comparison of different feature extraction methods for EEG-based emotion recognition | 2020 | [22] |
| Rahman, Md Asadur, et al | PCA Feature Selection – EEG Emotion | Employing PCA and t-statistical approach for feature extraction and classification of emotion from multichannel EEG signal | 2020 | [23] |
| Molcho, Lior, et al | Chi-Test Feature Selection – EEG Emotion | Single-channel EEG features reveal an association with cognitive decline in seniors performing auditory cognitive assessment. | 2022 | [24] |
| Caicedo-Acosta, Julian, et al | Lasso Feature Selection – EEG Emotion | Multiple-instance lasso regularization via embedded instance selection for emotion recognition | 2019 | [25] |

In this equation, PSO represents the Particle Swarm Optimization process. Loss is the loss function that PSO aims to minimize through hyperparameter optimization. X and Y represent the feature matrix and target vector, respectively, used in the XGBoost training process.

### C.  Combining PSO and Fuzzy Logic to Enhance XGBoost

When combining PSO and fuzzy logic to enhance XGBoost, the goal is to empower the strengths of both techniques. PSO optimizes the hyperparameters of the XGBoost model, including those that define the fuzzy system integrated into the model (if the fuzzification is parameter-dependent). The fuzzy logic provides a flexible framework for handling ambiguity and partial truths within the data, thereby potentially improving the model's robustness and accuracy. Meanwhile, PSO ensures that the model operates at its optimal parameter settings. This hybrid approach can result in a model that not only handles data with inherent uncertainties more effectively but also operates at an optimized level of performance across a range of conditions and datasets.

$$Enhanced\ Model\ = \\ XGBoost(PSO(\ FuzzyParams, Loss, X, Y)) \qquad (3)$$

Here, XGBoost is the base machine learning model. PSO denotes the Particle Swarm Optimization process used to optimize both the hyperparameters of the XGBoost model and parameters specific to the fuzzification process, collectively referred to as FuzzyParams. Loss represents the loss function optimized to achieve the best performance. X and Y are the input features and target outputs used in model training.

### D.  NGN Features

NGN feature selection on EEG signals in emotion recognition presents a sophisticated approach, especially when paired with an enhanced XGBoost classifier. This method excels at handling the spatially complex and high-dimensional nature of EEG data. Neural Gas Networks are unsupervised learning algorithms designed to preserve the topological properties of the input space, allowing them to identify complex patterns and relationships within the data that may be crucial for accurate emotion classification. The integration of these networks with an XGBoost classifier can strengthen these refined features to significantly boost the model's performance, enhancing its predictive accuracy and robustness in emotion recognition tasks.

$$Enhanced\ Feature\ Selected\ Model\ = \\ XGBoost(NGN\ (X, \eta, \lambda, \epsilon), Y) \qquad (4)$$

In which, NGN $(X, \eta, \lambda, \varepsilon)$ incorporates $NGN$ with: $X$ as the input EEG features. $\eta$ representing the learning rate parameter. $\lambda$ as the decay rate, affecting how the learning rate decreases over time. $\epsilon$ for the neighborhood function, influencing how the topological properties of the input space are preserved. XGBoost continues to function as the classifier, now using the feature set refined by NGN. Y are the labels for emotion classes.

## IV. PROPOSED METHOD

In our proposed methodology, we enhance the XGBoost classifier using an integration of Particle Swarm Optimization (PSO) and fuzzy logic tailored for emotion recognition through physiological signals. This hybrid approach starts with PSO, which optimizes the hyperparameters of XGBoost, such as learning rate, tree depth, and number of trees. Each particle in the PSO algorithm represents a potential set of hyperparameters, exploring the parameter space by following the swarm's best-known positions, thereby iteratively improving the model configuration. Concurrently, fuzzy logic is introduced to the XGBoost framework to manage the inherent uncertainties and imprecisions in the physiological data, allowing the model to handle ambiguous or overlapping emotional states more effectively. By fuzzifying the input features or decision boundaries, the model can make softer, more human-like decisions. This PSO-fuzzy enhanced XGBoost method significantly boosts the robustness and accuracy of emotion recognition by optimizing the model's parameters for best performance while effectively dealing with the complexities of physiological signal interpretation.

Also, the Neural Gas Network (NGN) plays a crucial role in the feature selection process for emotion recognition from physiological signals. NGN is an adaptive learning algorithm that organizes a set of neurons to optimally cover the input space without predefined connectivity rules, unlike traditional neural networks. The algorithm works by presenting input vectors sequentially to the network, where each neuron competes to be closest to the input vector. The winning neuron and its neighbors in the input space are then moved closer to the input vector, effectively learning the distribution and structure of the data. Over multiple iterations, this results in a self-organized representation of the input data where similar inputs are mapped close to each other, and dissimilar ones are farther apart. For feature selection, this property of NGN is exploited to extract the most salient features from the high-dimensional physiological data. These features are then used to train the XGBoost model, providing a robust and discriminative feature set that enhances the accuracy of emotion

Based on our research, it is the first time that PSO and fuzzy logic have been used to improve the XGBoost, especially in the field of emotion recognition and EEG signals. Also, upon investigation, no feature selection using NGN in the field of emotion recognition was found to be cited. The workflow of our proposed method is illustrated in Figure 1.

## V. VALIDATION AND RESULTS

### A. Dataset

We used a standard IEEE emotion recognition dataset called "BRAINWAVE EEG DATASET" [27, 28], which is available online at [26]. This dataset consists of brainwave EEG signals from eight subjects collected in a lab-controlled environment under a specific visualization experiment. The data include simple timestamps followed by the five bands of brainwave signals reading from the five electrodes of the emotive insight sensor: Theta, Alpha, Low Beta, High Beta, and Gamma. More than 10,000 brainwaves were collected. However, after applying several data filtering techniques, including the removal of noise signals and margins from the start and end of each picture showing time, only 1550 brainwaves remained. insider threats by analyzing brainwave activity through EEG signals involves understanding how certain brain activity patterns might correlate with deceptive or malicious intent, possibly identifying individuals who might pose an insider threat before any malicious actions occur. The concept involves using EEG to detect subconscious or conscious signs of malicious intent or deception in individuals. By analyzing brainwaves, researchers aim to find biomarkers or patterns that might indicate a risk of insider threats. Each picture is attached with two main values: Valence, which shows the degree of positive or negative effect the image evokes, and Arousal, which shows the intensity of the effect the image evokes. Images with a valence value equal to one are labeled as zero: ''High Risk''. Images with valence values equal to two and three are labeled as one: ''Medium Risk''. Images with valence values equal to four and five are labeled as two: ''Normal''. Images with valence values equal to six and seven are labeled as three: ''Low Risk''. All images selected as part of this experiment had arousal values of more than five to ensure their intense impact on the participants.

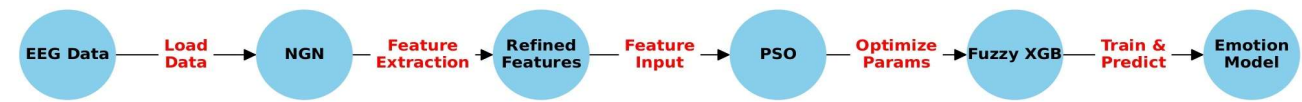

classification. This approach not only improves the efficiency of feature extraction but also contributes to the overall performance by reducing the dimensionality and focusing on the most informative aspects of physiological signals.

Fig. 1. Proposed Method's Workflow

### B. Classifiers

We employed five different classifiers for our evaluation comparison [29, 30]. Naive Bayes (NB) is a probabilistic classifier based on applying Bayes' theorem with strong (naive) independence assumptions between the features. It is particularly suited for high-dimensional data and is widely used in text and numerical classification due to its simplicity and effectiveness in dealing with large feature spaces. Logistic Regression (LR) is a statistical model that, in its basic form, uses a logistic function to model a binary dependent variable, although extensions to multiclass are available. It estimates the probability that an instance falls into a specific category, making it effective for binary classification problems. A Decision Tree is a non-parametric supervised learning method used for classification and regression tasks. The goal is to create a model that predicts the value of a target variable by learning simple decision rules inferred from the data features, resulting in a structure where each branch represents a choice between a number of alternatives, and each leaf node represents a classification or decision. XGBoost (Extreme Gradient Boosting) is an optimized distributed gradient boosting library designed to be highly efficient, flexible, and portable. It implements machine learning algorithms under the Gradient Boosting framework, providing a scalable, fast, and accurate method for regression, classification, and ranking problems. Fuzzy XGBoost incorporates fuzzy logic into the traditional XGBoost framework to handle datasets with uncertainty and imprecision effectively. By integrating fuzzy decision trees in the boosting process, this classifier allows for more nuanced thresholding and decision-making, accommodating data with overlapping or ambiguous boundaries between classes.

Table 2 compares classification results for different classifiers and feature selection using 70% for train and 30% for testing. All reported results belong to the test phase, and they were averaged over five runs for all four classes. Selected features are an average of 25%, 50%, and 75%. Also, all data is standardized before the classification process to ensure unified behavior for sensitive classifiers like LR.

## C. Feature Selection

The Chi-Square test [33] is a statistical method used to determine the independence of two categorical variables. In the context of feature selection, it is utilized to assess the relevance of each categorical feature in relation to the target variable. By performing a Chi-Square test between a feature and the target, one can determine whether the association between them is statistically significant. Features that show a strong association—indicated by a low p-value in the Chi-Square test—are considered relevant and selected for use in the model. This method is particularly useful for filtering out features that do not have a significant relationship with the target variable, thereby simplifying the model and potentially improving its performance by focusing on more relevant inputs. Principal Component Analysis (PCA) [31] is a statistical technique used for dimensionality reduction while preserving as much variability as possible. It works by identifying the directions, called principal components, along which the variance of the data is maximized. In feature selection, PCA is used to transform the original correlated features into a new set of uncorrelated variables. Each principal component is a combination of the original features, with the first few components capturing the majority of the variance in the data. By selecting a subset of these components, typically those that account for a significant portion of the variance, you can effectively reduce the feature space. This not only helps simplify the model but also alleviates issues like multicollinearity and overfitting, making the model more generalizable. Least Absolute Shrinkage and Selection Operator (LASSO) [32] is a regression analysis method that performs both variable selection and regularization in order to enhance the prediction accuracy and interpretability of the statistical model it produces. Unlike traditional regression methods that minimize the sum of squared residuals, Lasso aims to minimize a combination of this sum and the sum of the absolute values of the coefficients multiplied by a tuning parameter, $\lambda$. This addition of the penalty term encourages the solution to have fewer non-zero coefficients, effectively reducing the number of features in the model. By adjusting $\lambda$, Lasso can be used to select only the most important features, discarding those that contribute little to the model's predictive power. This makes Lasso particularly useful for feature selection in scenarios with high dimensionality or when feature reduction is crucial. Figure 2 depicts 15 features selected from the NGN dataset.

## D. Fuzzy Logic and PSO Parameters

The features are fuzzified based on quantile thresholds. Specifically, the lower threshold is set at the 33rd percentile, and the upper threshold is at the 67th percentile of the feature values. This divides the data into three fuzzy states: low (0), medium (1), and high (2). These thresholds help in categorizing the feature values into three distinct groups, allowing the model to handle the data in a way that mimics human decision-making processes. The objective function defined for the PSO algorithm is to minimize the loss, which is computed as 1 - accuracy from an XGBoost classifier model. The parameters optimized using PSO are max_depth (range from 3 to 10) and learning_rate (range from 0.01 to 0.3). These are the hyperparameters of the XGBoost model, which significantly affect model performance. The lower bound for max_depth is set at 3 and for learning_rate at 0.01. The upper bounds are 10 for max_depth and 0.3 for learning_rate. The PSO algorithm is executed with a swarm size of 50 and a maximum of 100 iterations, aiming to explore a broad range of the parameter space to find the optimal settings that maximize the accuracy of the XGBoost model. PSO's best cost over iterations for the PSO fuzzy XGBoost experiment is presented in Figure 3. Also, Figure 4 illustrates the fuzzy membership functions for one of the selected features. Figure 5 shows the Test phase confusion matrixes of XGB and PSO Fuzzy XGB.

TABLE II.    CLASSIFICATION RESULTS

| Classifier | Metric | Chi-t | PCA | Lasso | NGN | Raw |
|---|---|---|---|---|---|---|
| **NB** | *Avg* | 0.343 | 0.346 | 0.345 | 0.365 | **0.396** |
| | *STD* | **0.009** | 0.025 | 0.029 | **0.009** | 0.014 |
| **LR** | *Avg* | 0.293 | 0.374 | 0.351 | 0.375 | **0.431** |
| | *STD* | **0.015** | 0.018 | 0.036 | 0.028 | 0.019 |
| **DT** | *Avg* | 0.763 | 0.606 | 0.796 | 0.840 | **0.852** |
| | *STD* | 0.016 | 0.031 | 0.025 | **0.015** | 0.026 |
| **XGB** | *Avg* | 0.764 | 0.608 | 0.802 | **0.904** | 0.879 |
| | *STD* | 0.012 | 0.027 | 0.028 | **0.010** | 0.024 |
| **Fuzzy XGB** | *Avg* | 0.868 | 0.634 | 0.551 | 0.851 | **0.890** |
| | *STD* | 0.013 | 0.018 | 0.017 | **0.009** | 0.010 |
| **PSO Fuzzy XGB** | *Avg* | 0.872 | 0.763 | 0.782 | **0.934** | 0.915 |
| | *STD* | 0.025 | 0.050 | 0.032 | 0.031 | **0.008** |

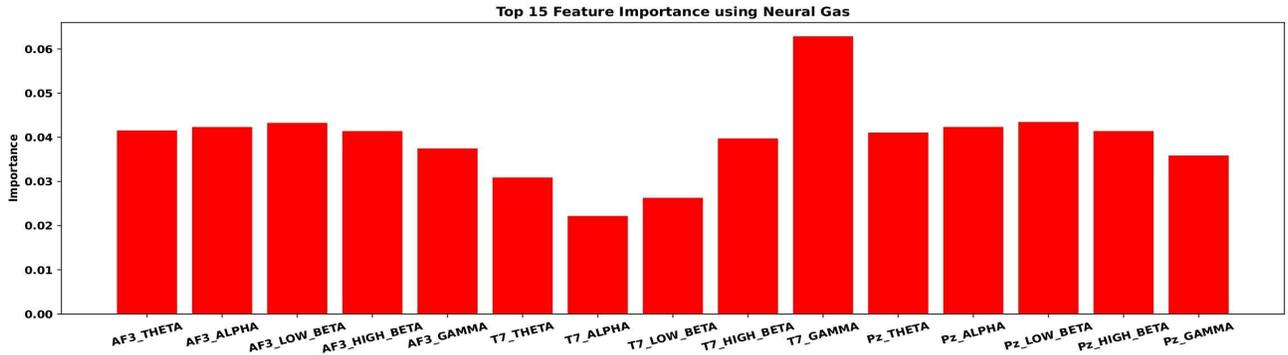

Fig. 2. The Most impactful Features Selected by NGN

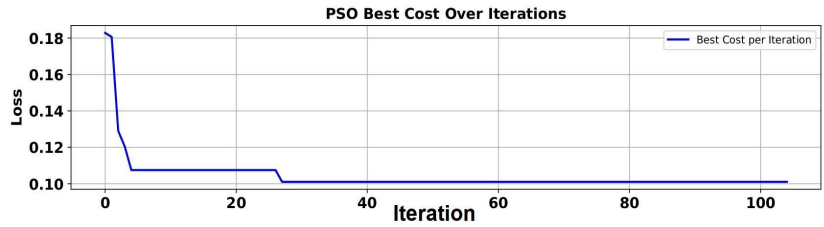

Fig. 3. PSO's best cost for PSO Fuzzy XGBoost Experiment

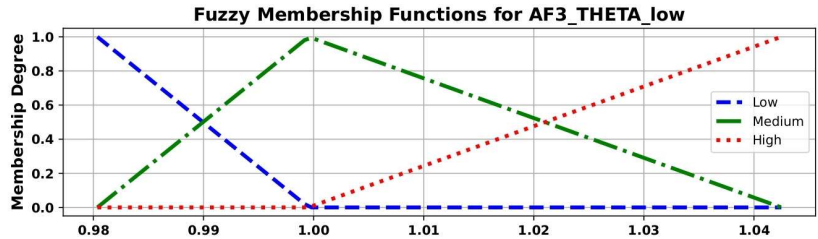

Fig. 4. Fuzzy membership functions of one of the selected features

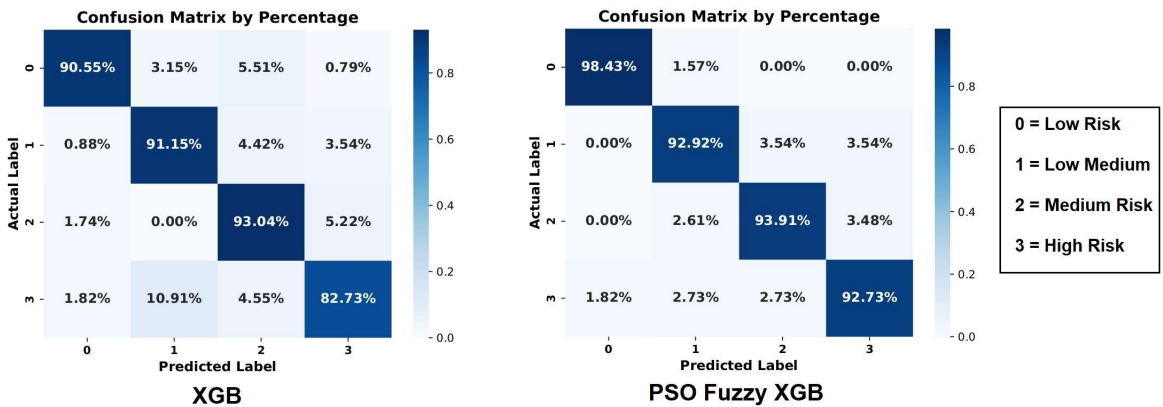

Fig. 5. Test Phase Confusion Matrixes of XGB and PSO Fuzzy XGB

## E. Discussion

For Naïve Bayes (NB), the performance is relatively stable across feature selection methods, with a slight superiority when no feature selection (Raw) is applied. Logistic Regression (LR) shows improved performance with Principal Component Analysis (PCA), Lasso, and Neural-Guided Network (NGN) compared to Chi-squared (Chi-t), with the highest performance using the Raw data. Decision Trees (DT) and both versions of XGBoost (XGB and Fuzzy XGB) show a clear advantage when using NGN and Raw data, indicating a potential for these methods to handle complex models effectively. Notably, PSO Fuzzy XGB exhibits exceptional performance with NGN and Raw data, achieving the highest average scores while also maintaining competitive performance across other methods despite higher variability, as indicated by the standard deviations. This suggests that PSO Fuzzy XGB, especially with NGN, could be particularly effective in scenarios where robust feature extraction and high prediction accuracy are critical.

## VI. CONCLUSIION

In conclusion, our study has made significant strides in advancing the field of emotion recognition using physiological signals through the integration of Neural-Gas Network (NGN), XGBoost, Particle Swarm Optimization (PSO), and fuzzy logic. Our results affirm that the PSO-fuzzy XGBoost classifier not only enhances the accuracy of emotion recognition systems but also outperforms traditional classifiers and feature selection techniques. NGN's ability to effectively adapt to complex input spaces without predefined structures proves crucial in handling high-dimensional data and improving feature extraction, which is fundamental in physiological signal analysis. Furthermore, the incorporation of fuzzy logic facilitates the management of fuzzy data and introduces human-like reasoning capabilities into our models' decision-making processes. For future work, several avenues could be explored to further enhance emotion recognition systems. First, experimenting with larger and more varied datasets could help validate and refine the models, ensuring their robustness and applicability across different demographic and psychological profiles. Second, integrating multimodal data sources, such as combining physiological signals with facial expressions and voice patterns, could provide a more holistic approach to emotion recognition. Additionally, exploring other optimization algorithms alongside or in place of PSO could uncover more efficient ways to tune hyperparameters and optimize model performance. Lastly, delving deeper into the theoretical underpinnings of NGN and its potential synergies with other machine-learning approaches could unlock new methodologies for feature selection and classification in complex datasets. Through continued research and development, these enhancements could lead to more sophisticated, accurate, and accessible emotion recognition technologies, thereby expanding their practical applications in fields such as healthcare, marketing, automotive industries, and beyond.